# Evaluating Large Language Models in Analysing Classroom Dialogue：a Case Study

Yun Long[1#], Haifeng Luo[1#], Yu Zhang[1*]

[1] Institute of Education, Tsinghua University, Beijing, China, 100084

[*] Corresponding coauthor: Yu Zhang, zhangyu2011@tsinghua.edu.cn +86 10 62785686 (Tel)

[#] First coauthors

## ABSTRACT

This study explores the application of Large Language Models (LLMs), specifically GPT-4, in the analysis of classroom dialogue, a crucial research task for both teaching diagnosis and quality improvement. Recognizing the knowledge-intensive and labor-intensive nature of traditional qualitative methods in educational research, this study investigates the potential of LLM to streamline and enhance the analysis process. The study involves datasets from a middle school, encompassing classroom dialogues across mathematics and Chinese classes. These dialogues were manually coded by educational experts and then analyzed using a customised GPT-4 model. This study focuses on comparing manual annotations with the outputs of GPT-4 to evaluate its efficacy in analyzing educational dialogues. Time efficiency, inter-coder agreement, and inter-coder reliability between human coders and GPT-4 are evaluated. Results indicate substantial time savings with GPT-4, and a high degree of consistency in coding between the model and human coders, with some discrepancies in specific codes. These findings highlight the strong potential of LLM in teaching evaluation and facilitation.

KEYWORDS: Artificial Intelligence; Large Language Models; ChatGPT; education; classroom dialogue analysis

## 1. INTRODUCTION

### 1.1 Classroom dialogue and its impact

Learning happens during human interactions (Mercer, 2013), and classroom dialogue is one of the fundamental ways to initiate such interactions in formal schooling according to social-cultural perspectives,. Compared with other modal of interactions, language plays a significant role in the development of thinking. Classroom dialogue has thus been an essential focus of academic inquiry for decades, recognized as pivotal in the process of creating meaning and thereby fundamental to the educational experience (Mortimer & Scott, 2003, p. 3). Empirical studies have revealed the benefits of teaching dialogic skills to students, noting enhancements in their reasoning abilities and collaborative problem-solving capabilities (Howe & Abedin 2013; Kuhn 2015, 2016; Mercer, 2013; Mercer et al., 2004). Dialogic pedagogy, inherently participatory, aims to validate and expand all participants' contributions in classroom dialogues (Nystrand et al. 2003). Its primary goal is to cultivate learner autonomy, encouraging students to collaboratively seek understanding, build upon their ideas, and integrate others' views to refine their thought processes (Flitton & Warwick, 2012, p.3). Evidence also points to academic improvements in various subjects as a direct outcome of dialogic pedagogical approaches (Baines et al., 2007; Mercer et al., 2003; Rojas-Drummond et al., 2010; Rojas-Drummond & Mercer, 2003). Notably, a UK-based efficacy trial involving nearly 5000 students reported moderately robust results (EEF, 2017). This trial, focusing on Dialogic Teaching for children aged 9–10, demonstrated a positive impact on their performance in English, science, and mathematics, roughly equating to an additional two months of academic progress.

## 1.2 The challenges in classroom dialogue analysis

Despite the established qualitative methods traditionally used in analyzing classroom dialogue, such as content, discourse, and thematic analysis, challenges remain in the efficiency and objectivity of these approaches. The typical workflow of dialogue analysis involves transcription, coding, and interpretation. Transcription, the conversion of spoken words in recorded classrooms into text, has been expedited by software, significantly reducing time and human resource demands. However, the subsequent coding process presents a more formidable hurdle, both in terms of complexity and time expenditure.

Coding, the act of sorting and structuring data into categories based on themes, concepts, or emergent significant elements, is indispensable in qualitative research. This method enables a systematic dissection of voluminous data, such as interview transcripts or observational records, which defy straightforward quantification. The coding process entails delving into the data to unearth distinct concepts and patterns, ultimately leading to the formation of themes that encapsulate the core of the data. This aspect of qualitative research, especially in educational domains where data can be particularly intricate, is marked by an iterative and adaptable approach. Unlike its quantitative counterpart, which aims to numerically represent data, qualitative coding embraces the researcher's subjectivity and the contextual nuance of the data.

This subjective dimension, however, introduces potential complications in educational research. The researcher's inherent biases and viewpoints may color their interpretation, potentially leading to skewed outcomes. Furthermore, the labor-intensive and resource-heavy nature of coding in qualitative educational research poses challenges, particularly when handling extensive datasets or conducting long-term studies. Ensuring the reliability and validity of coded qualitative data, especially in the diverse and variable contexts of educational research, remains a complex and demanding endeavor.

## 1.3 The potential role of large language models in educational research

In the ever-evolving domain of educational technology, the integration of Artificial Intelligence (AI) has marked a significant milestone in shaping learning outcomes. At the vanguard of these AI advancements are Large Language Models (LLMs), which epitomize the cutting-edge capabilities in natural language processing (NLP), a key component in educational interactions (Brown et al., 2021). The profound ability of LLMs to process and interpret verbal exchanges in educational settings has been recently underscored by Liu and colleagues (2023), highlighting their potential in transforming traditional educational methodologies.

LLMs, such as GPT-4, have demonstrated their capacity to detect patterns and trends within educational interactions, categorizing and synthesizing dialogue data effectively (Kumar, 2022). This ability paves the way for large-scale qualitative analysis, a feat previously daunting due to the limitations of conventional research methods. The current study is thus poised to evaluate how generative AI can be strategically employed in the educational field, leveraging these advanced models to gain deeper insights into learning processes.

With a particular focus on classroom dialogue, this research seeks to explore the role of LLMs in understanding and enhancing communication and interaction within educational contexts. The choice of classroom dialogue as a case study is deliberate, considering the centrality of communication in the learning experience. By harnessing the capabilities of LLMs, this study aims to offer a novel perspective on the analysis of educational discourse, contributing to the broader narrative of how AI, particularly generative AI, can revolutionize educational research and practice. This exploration not only aligns with the contemporary trend of digital integration in education but also opens new avenues for personalized and effective learning strategies, tailored to the unique needs of diverse educational landscapes.

## 2. BACKGROUND AND RELATED WORK

### 2.1 Large Language Models

Large language models (LLMs) represent a class of highly sophisticated AI tools primarily designed for tasks involving natural language processing (Brown et al., 2020; Radford et al., 2019). These models have revolutionized the approach to understanding and generating human language through advanced computational methods.

There are few fundamental capabilities of LLMs. First of all, LLMs are adept at comprehending complex language constructs, enabling them to grasp context, infer meaning, and understand subtleties and nuances in language (Bommasani et al., 2021). Their proficiency extends to understanding idiomatic expressions and detecting sentiment, making them versatile in various linguistic analyses. Secondly, these models excel in generating coherent, contextually appropriate, and often creative text outputs. They are used in applications ranging from content creation to conversation agents (Radford et al., 2019). Their ability to produce text that closely mimics human writing has significant implications for fields like journalism, creative writing, and customer service. Last but not least, LLMs are known for their exceptional pattern recognition abilities. They can identify and learn from patterns in data, a capability that is fundamental to tasks such as language translation, summarization, and even predictive text generation (Brown et al., 2020). In general, LLMs' fundamental capabilities including natural language understanding, natural language generation and pattern recognition.

In terms of the mechanism of LLMs, they are trained on extensive and diverse datasets, which include a wide array of text sources like books, articles, and websites (Bender et al., 2021). This training involves processing billions of words and requires sophisticated algorithms to manage and learn from such a vast amount of data. The training of LLMs involves deep learning techniques, particularly transformer models, which have shown remarkable efficiency in handling sequential data like text (Vaswani et al., 2017). These models learn to predict and generate language by identifying patterns and relationships within the training data. The diverse training allows these models to handle a wide range of language tasks and adapt to different styles, dialects, and jargon. This adaptability is crucial for applications that require a nuanced understanding of language (Devlin et al., 2018).

Based on the introduced capabilities and mechanism of LLMs, a standout feature of LLMs is their ability to contextualize information, enabling them to provide more accurate and relevant responses in conversational AI or content generation (Bommasani et al., 2021). Therefore, the advent of LLMs like GPT-4 signifies a major leap forward in AI's language capabilities. They offer transformative potential across numerous sectors, from education and customer service to content creation and beyond (Brown et al., 2020).

**2.2 GPT-4（GPTs）**

GPT-4, or Generative Pre-trained Transformer 4, is the latest iteration in the GPT series developed by OpenAI. It represents a significant leap forward in the capabilities of large language models (LLMs) for natural language processing tasks. Building on the advancements of its predecessors, GPT-4 is distinguished by its larger size, improved training algorithms, and enhanced capacity to understand context and generate human-like text (OpenAI, 2023). Additionally, the model finds diverse applications ranging from creating content, answering queries, language translation, and even coding, showcasing a remarkable breadth of knowledge and adaptability.

In recent advancements within the realm of artificial intelligence, OpenAI has introduced a novel and innovative concept titled "My GPTs," a significant extension of their existing Generative Pre-trained Transformer (GPT) series. This new development, as outlined by OpenAI (2023), represents a transformative step in democratizing AI technology by enabling users to create and customize their own versions of the ChatGPT model for specific applications, ranging from everyday tasks to specialized professional uses. The core of "My GPTs" lies in its user-friendly interface, which allows individuals to build personalized AI models without any prerequisite coding skills. This feature not only broadens the accessibility of AI technology but also fosters a more inclusive approach towards its development and utilization.

The "My GPTs" concept aligns with OpenAI's broader vision of integrating AI into real-world applications. Developers are provided with the capability to define custom actions for their GPTs, enabling these models to interact with external data sources or APIs, thereby enhancing their real-world applicability. This advancement signifies a major leap in the evolution of AI as 'agents' capable of performing tangible tasks within various domains (OpenAI, 2023).

In conclusion, OpenAI's "My GPTs" marks a significant milestone in the journey towards accessible, customizable, and safe AI, fostering a community-driven approach to AI development and application. This initiative not only expands the horizons of AI technology but also invites a diverse range of users to actively participate in shaping the future of AI, resonating with OpenAI's mission to build AI that is beneficial to humanity (OpenAI, 2023). In this study, a GPTs is developed based on the coding rules for analysing classroom dialogue.

**2.3 Classroom dialogue analysis framework**

Classroom dialogue has been heavily researched in recent years due to its perceived role in student learning (e.g., Alexander, 2008; Howe & Abedin, 2013; Mercer & Dawes, 2014; Schwarz & Baker, 2016). Influenced by socio-cultural perspectives, authors in this field view learning as a social activity, mediated through dialogue. Specifically, dialogue is perceived as the intermediary between collective and individual thinking (Vygotsky, 1962). Its quality, therefore, becomes particularly important as it determines the quality of collective thinking and, through this, individual progress. These views have resulted in research which aims to identify forms of dialogue that promote higher order thinking and, thus, are optimal for learning. Thanks to this research, there is now a fair degree of consensus over which forms are especially productive (Littleton & Mercer 2013). The characteristics of optimal classroom dialogue proposed by Alexander (2008) have proved particularly influential. According to Alexander, classroom dialogue should be: 1) collective with participants reaching shared understanding of a task; 2) reciprocal with ideas shared among participants; 3) supportive with participants encouraging each other to contribute and valuing all contributions; 4) cumulative, guiding participants towards extending and establishing links within their understanding; and 5) purposeful, that is directed towards specific goals. Similar forms of dialogue have been highlighted in the context of student–student interaction. Littleton and Mercer (2013) have identified three types of student–student talk: disputational, cumulative and exploratory. Characterised by disagreement and individualised decisions, disputational talk was thought to be the least educationally productive. Some educational value was attributed to cumulative talk, as it was characterised by general acceptance of ideas, but lack of critical evaluation. Exploratory talk was observed less frequently; yet, it was regarded as the most educationally effective. It involved participants engaging critically with ideas and attempting to reach consensus. Initiatives, like the 'Thinking Together' programme (Dawes, Mercer, and Wegerif, 2003; Mercer and Littleton 2007), aimed to promote primary school children's use of exploratory talk, and showed a positive impact on students' problem solving, mathematics and science attainment/learning. Likewise, 'accountable talk' has been promoted as the most academically productive classroom talk (Michaels, O'Connor, and Resnick 2008). It encompasses accountability to: 1) the learning community, through listening to others, building on their ideas and expanding propositions; 2) accepted standards of reasoning (RE), through emphasis on connections and reasonable conclusions; and 3) knowledge, with talk that is based on facts, texts or other publicly accessible information and challenged when there is lack of such evidence.

Working in secondary classrooms, Nystrand et al. (1997) characterised dialogic instruction via three key discourse moves that teachers might make: 1) authentic questions, which are questions with no predetermined answers; 2) uptake, which occurs when previous answers are incorporated into subsequent questions; and 3) high-level evaluation, which occurs when teachers elaborate or ask follow-up questions in response to students' replies, instead of giving a simple evaluation, such as 'Good' or 'OK' (Nystrand et al. 2003). While there are differences between these approaches, there are also marked commonalities, regardless of whether the research refers to whole class or small group contexts. Shared features include:

- invitations that provoke thoughtful responses (e.g. authentic questions, asking for clarifications and explanations);
- extended contributions that may include justifications and explanations;
- critical engagement with ideas, challenging and building on them;
- links and connections;
- attempts to reach consensus by resolving discrepancies.

For these features to occur, a generally participative ethos is important, with participants respecting and listening to all ideas. This necessitates making the discourse norms accessible to all (Michaels, O'Connor, & Resnick 2008). Changing the classroom culture in this manner might be a challenge for any teacher. For better analysing classroom dialogues, researchers proposed various coding schemes, in this study, we employed the coding scheme conducted by Cambridge Educational Dialogue Research Group, and revised it considering the context of Chinese classrooms. Table 1 shows the codes and their brief definitions.

Table 1. The coding scheme used for analysing classroom dialogue (revised from Cambridge Dialogue Analysis Scheme, Howe et al., 2019)

| Codes | Brief definitions (and key words) |
|---|---|
| **Elaboration invitation (ELI)** | Invites building on, elaboration, evaluation, clarification of own or another's contribution. |
| **Elaboration (EL)** | Builds on, elaborates, evaluates, clarifies own or other's contribution (if own, it should be on separate turns) within an exchange. This adds substantive new information or a new perspective beyond anything said in previous turns, even by one word. |
| **Reasoning invitation (REI)** | Explicitly invites explanation/justification of a contribution or speculation (new scenarios) /prediction/hypothesis. |
| **Reasoning (RE)** | Provides an explanation or justification of own or another's contribution. Includes drawing on evidence (e.g. identifying language from a text/poem that illustrates something), drawing analogies (and giving reasons for them), making distinctions, breaking down or categorising ideas. |
| **Co-ordination invitation (CI)** | Invites synthesis, summary, comparison, evaluation or resolution based on two or more contributions (i.e. invites all descriptors of SC and RC below) |
| **Simple co-ordination (SC)** | Synthesises or summarises collective ideas (at least two, including own and/or others' ideas). Compares or evaluates different opinions, perspectives and beliefs. Proposes a resolution or consensus view after discussion. |
| **Reasoned co-ordination (RC)** | Compares, evaluates, resolves two or more contributions in a reasoned fashion (e.g. 'I agree with Susan because her idea has more evidence behind it than Emma's'). |
| **Agreement (A)** | Explicit acceptance of or agreement with a statement(s) (e.g. 'Brilliant', 'Good', 'Yeah', 'Okay', I agree with X...). |
| **Querying (Q)** | Doubting, full/partial disagreement, challenging or rejecting a statement. Challenging should be evident through verbal means. |
| **Reference back (RB)** | Introduces reference to previous knowledge, beliefs, experiences or contributions (includes procedural references) that are common to the current conversation participants. This should refer to a specific activity or time point, not just simple recall (e.g. 'Do you remember what we call it?'). |
| **Reference to wider context (RW)** | Making links between what is being learned and a wider context by introducing knowledge, beliefs, experiences or contributions from outside of the subject being taught, classroom or school. |
| **Structural silence (SU)** | Students may feel 'silenced', such type of silence may be linked to social situations and interpersonal interactions. |
| **Strategic silence (SA)** | Students choose not to express or articulate an utterance. This experience of strategic silence may be more personal in nature, and although the motivation to self-silence may be influenced by the interactions of others, the decision to remain silent remains at a more private level. |
| **Other Invitation (OI)** | Invitations cannot be coded as any code related to invitation provided above. |
| **Other (O)** | Dialogue turns cannot be coded as any code provided above, |

Specifically, in this coding scheme, the elaboration invitations (ELI) and reasoning invitations (REI) categories captured authentic questions that provoked thoughtful answers (e.g. Nystrand et al. 1997). The elaboration (EL), RE and Querying (Q) categories captured core features of exploratory talk (e.g. Littleton & Mercer, 2013) and accountable talk (Michaels, O'Connor & Resnick, 2008); namely building on ideas, justifying and challenging, respectively. The co-ordination invitations (CI) category addressed invitations to synthesis ideas, while simple co-ordination (SC) and reasoned coordination (RC) addressed responses to such invitations, the difference between RC and SC being that RC draws on evidence, theory or a mechanism for justification (Felton & Kuhn 2001; Osborne et al. 2004). Establishing links and identifying connections, stressed by Alexander (2008) and Michaels, O'Connor, and Resnick (2008), were represented by the reference back (RB) and reference to wider context (RW) categories, which focus respectively on prior knowledge or beliefs and the wider context. Two further codes are not directly mappable onto current conceptions of productive dialogue: agreement (A) and other Invitations (OI). Nevertheless, in combination with ELI or EL, A represents high-level evaluation (Nystrand et al., 2003)). Nystrand et al. (2003) highlight 'simple evaluation plus elaboration' and 'simple evaluation plus follow-up question' as high-level teacher evaluations of student responses. In our coding system, the first example is captured through the combination of A and EL and the second through the combination of A and ELI. As for OI, this category was included to contrast ELI, REI and CI with less productive invitations.

## 3. DATA

In this study, we harness diverse datasets to evaluate the efficacy and applicability of AI, particularly GPT-4, in educational settings. Our data sources include dialogue transcripts in Chinese classes and manual and automated coding results. These data contributing insights into the potential of AI in enhancing educational experiences.

### 3.1 Classroom dialogue transcripts

The first dataset comprises classroom dialogue transcripts, which are crucial in understanding real-world educational interactions. These classroom dialogues are sourced from a high school in Beijing and encompass two main subjects and educational levels. The classroom dialogue recordings were automatically transcribed by a software named Feishu Miaoji into texts and then manually checked by research assistants. The richness and diversity of these texts allow for a comprehensive analysis of how AI can be integrated into classroom dialogues effectively.

### 3.2 Manual coding data

To ensure the quality and relevance of our analysis, a significant portion of our data has undergone meticulous manual coding. This dataset includes classroom dialogues and interactions that have been annotated by education professionals. Annotations has been done based on the coding scheme and rules. This manual coding serves as a benchmark for assessing the accuracy and contextual appropriateness of AI-generated responses and analyses.

### 3.3 GPT-4 coding data

The third dataset involves the application of GPT-4 and the costumised GPTs for coding and analyzing classroom dialogues. This cutting-edge AI model offers insights into the potential of machine learning in educational contexts. We specifically focus on how GPT-4 interprets educational dialogues, its effectiveness in identifying key educational elements, and its capability to generate meaningful and contextually relevant responses. By comparing the GPT-4 coding with manual annotations, we aim to evaluate the practicality and reliability of AI in educational settings, potentially paving the way for AI-assisted educational tools.

## 4. EXPERIMENT

### 4.1 Participants

The study involved students from two classes, one each from the first and second grades of junior high school, at a middle school in Beijing. Along with these students, their mathematics and Chinese language teachers also participated in the experiment. This selection of participants was strategic, aiming to provide a diverse range of perspectives and experiences in the classroom setting.

In both the mathematics and Chinese language subjects, six lessons from each were meticulously selected for this study. These lessons were chosen based on their representativeness of the typical curriculum and their potential to generate rich, meaningful dialogue for analysis. The aim was to ensure a broad spectrum of classroom interactions were captured, encompassing various teaching styles, student responses, and interactive dialogues.

For each of these subjects, the chosen lessons underwent a detailed process of annotation. This process was twofold: one, a manual annotation performed by experts in classroom dialogues, and two, an automatic annotation carried out by the customized GPTs specifically designed for analyzing classroom dialogues. The manual annotation served as a benchmark, a standard against which the GPTs' performance could be measured.

The volume of data collected for this experiment was substantial, exceeding 150,000 characters. This extensive dataset was crucial for providing a comprehensive analysis. The length of the data ensured that a wide range of classroom interactions were included, allowing for a more thorough and nuanced examination of the AI's capabilities in interpreting and analyzing educational dialogues.

The inclusion of both mathematics and Chinese language subjects was particularly significant. It allowed the experiment to explore the AI's versatility in handling different types of academic content – the numerical and problem-solving focus of mathematics and the linguistic and interpretative nature of Chinese language. This diversity in subject matter was expected to provide deeper insights into the AI's adaptability and effectiveness across varying academic disciplines.

Overall, the participation of students and teachers from these specific classes, along with the careful selection of lessons and extensive data collection, set the stage for a robust and insightful exploration into the use of AI in analyzing educational dialogues.

4.2 Procedures

Figure 1 presents the technological roadmap utilized in our study, illustrating a comprehensive workflow from initial data acquisition to the final comparative analysis. The flowchart commences with the collection of classroom video recordings, which are then transcribed into text. Both manual and GPT-driven annotations are applied following the same predetermined coding scheme, with the manual process serving as a benchmark for the AI's performance. The GPT model, tailored to these specific educational analysis tasks, annotates the dialogues autonomously, paralleling the manual effort. The culmination of this methodology is a critical comparative analysis, juxtaposing the manual annotations with those generated by the GPT model. The roadmap encapsulates the systematic approach adopted in our study, reflecting the meticulous integration of AI tools like GPT-4 with conventional methods to evaluate and enhance the analysis of educational dialogues.

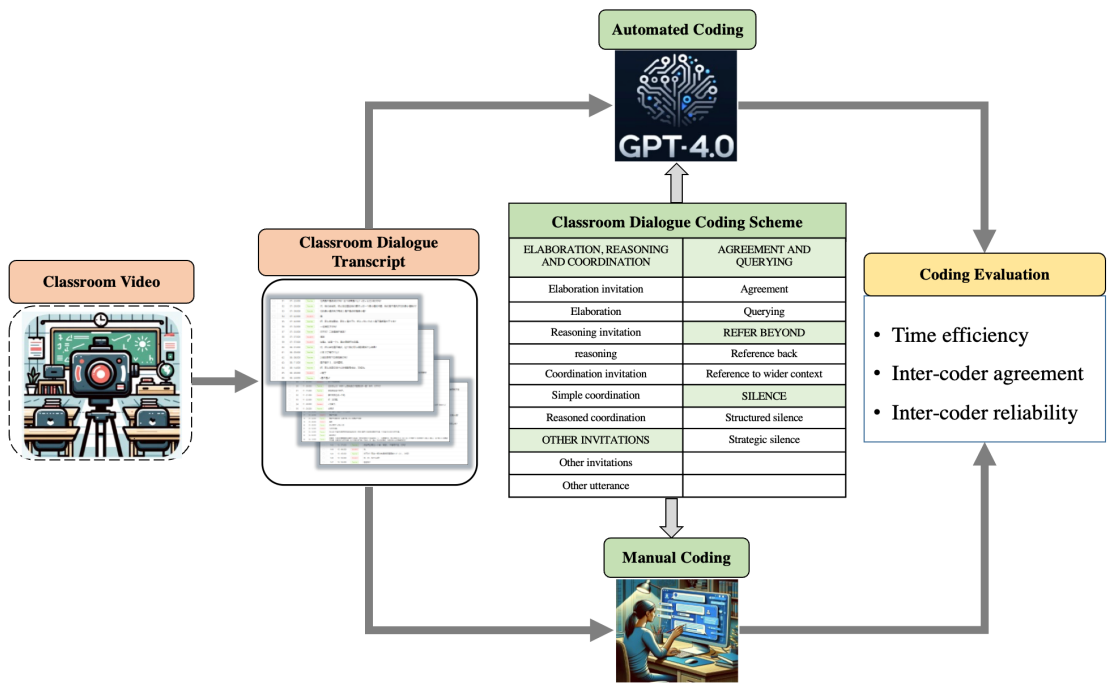

Figure 1. Technology roadmap

4.2.1 Transcription

The video tapes of classroom dialogues need to be transcript into text for analysis. Following a thorough evaluation of various automated transcription solutions, Feishu

Miaoji was chosen, attributing to its exemplary features. The standout capability of Feishu Miaoji is its provision of time stamps, a critical attribute for aligning the transcribed text with precise instances within the video recordings. This functionality significantly enhances the ease of playback and systematic organization of the data. Furthermore, Feishu Miaoji supports the identification and attribution of dialogue to specific speakers and offers facilities for post-transcription modifications, which proved immensely beneficial for our research objectives. The granularity and precision of transcription afforded by Feishu Miaoji were indispensable for the accurate analysis and subsequent interpretation of the interactions observed in the classroom settings.

4.2.2 Organising transcripts

Once the transcription was complete, the next step involved meticulously organizing the transcribed text. We manually transferred the transcript into Excel spreadsheets for better data management. In these spreadsheets, each dialogue turn was separately identified and formatted uniformly. This organization was critical to ensure consistency and ease of access for subsequent analysis stages. The structured format helped in segregating and identifying distinct conversational elements, laying a clear foundation for both manual and AI-driven annotations.

4.2.3 Manual Coding

The manual coding process was conducted based on a predefined coding scheme (see Table 1). Every dialogue turn in the classroom conversations was encoded by educational experts. This meticulous process involved scrutinizing each dialogue turn to categorize and code them according to the established coding rules. The manual coding served as a gold standard, offering a comprehensive and nuanced understanding of the classroom dynamics.

4.2.4 GPT Annotation

Parallel to the manual coding, we employed a trained GPT model to annotate the classroom dialogues. The GPT model was programmed to follow the same coding scheme used in the manual process. We put the classroom dialogue text into the GPT system, which then autonomously coded each dialogue turn. This step was crucial in evaluating the proficiency of AI in understanding and interpreting classroom interactions as compared to human experts.

4.2.5 Comparison

The final step involved a comparative analysis between the manual coding and those results generated by the GPT model. Three important dimensions are considered in the evaluation: efficiency, inter-coder agreement, and inter-coder reliability. Time efficiency in this context refers to the duration required to complete the coding process, comparing the speed of human coders against the GPT-based automatic coding system. The operational algorithm includes task definition, measure duration, repeat measurement, and analyse data. While there is no complex formula, it can be represented simply as :

$$Time\ Efficiency = \frac{Time\ Taken\ to\ Coding\ by\ Human\ Coder}{Time\ Taken\ to\ Coding\ by\ GPT}$$

Inter-coder agreement here refers to the degree of agreement or consistency between human coders and the GPT-based coding on the same dataset. The operational algorithm includes standardize coding scheme, independent coding, collect and compare codes, caculate agreement ratio. The percentage agreement is calculated as:

$$Percentage\ Agereement = \left(\frac{Number\ of\ AGreements}{Total\ Number\ of\ Decisions}\right) \times 100\%$$

Inter-coder reliability in this context measures the consistency of coding outcomes between human coders and the GPT system across different datasets or coding instances, taking into account the possibility of chance agreement. The operational algorithm includes ensure consistent coding scheme, perform independent coding, calculate reliability coefficients, and analyse and interpret results. Inter-coder reliability, particularly when comparing human to automated coding, can be assessed using Cohen's Kappa (κ), and the formula is shown below:

$$k = \frac{P_o - P_e}{1 - P_e}$$

*Here Po is the relative observed agreement among raters, and Pe is the hypothetical probability of chance agreement.

4.3 Data analysis

A comprehensive examination was conducted focusing on three distinct yet interconnected aspects: the assessment of time savings provided by using GPTs for classroom dialogue coding, the inter-coder agreement percentage and inter-coder reliability between human coders and ChatGPT.

The time efficiency brought about by the use of GPTs in coding was a critical aspect of our analysis. By comparing the duration for manual coding tasks with those completed using GPTs, we could discern the practical benefits of AI in educational research. This aspect is pivotal, particularly in academic settings where efficient time management directly translates to enhanced productivity and resource optimization. The analysis revealed significant time savings when employing GPTs, suggesting a substantial impact on operational efficiency in educational research settings.

We then analysed the inter-coder agreement. This step is crucial for validating the reliability of AI-assisted coding against human standards. We computed the percentage of agreement and the inter-coder reliability using Cohen's Kappa statistic, which is a widely recognized measure in research for its ability to account for chance agreement. This metric was particularly important to gauge the level of consistency between human coders and ChatGPT, ensuring that the coding of AI aligns accurately with the predefined coding manual. The use of SPSS software enabled a thorough cross-tabulation, facilitating a nuanced understanding of how well ChatGPT's coding decisions matched with those of human coders.

In summary, our data analysis provided a two-faceted evaluation of AI's role in coding classroom dialogue. It highlighted not only the technological efficiency of AI in mirroring human coding accuracy but also underscored the practical time-saving benefits. This comprehensive approach ensures a balanced assessment, catering to both the methodological rigor required in qualitative research and the pragmatic needs of educational practitioners.

**4.4 Results**

**4.4.1 Time efficiency evaluation**

For the analysed Chinese lessons, the cumulative duration was approximately 4 hours and 7 minutes, while for the Math lessons it was about 5 hours and 17 minutes. Utilizing ChatGPT, and excluding instances where dialogue turns exceeded the input limitations for the AI, the total time spent on analysis did not exceed one hour. To calculate the time savings compared to manual coding, we selected a single Math lesson for a timed comparison. This lesson, lasting 41 minutes and 29 seconds with 82 dialogue turns, was processed by ChatGPT in increments of 10 dialogue turns to avoid skipping, totaling 5 minutes of analysis time. The manual coding process involved an experienced researcher watching the lesson video at 1.5x speed, reviewing the transcript twice, and coding each dialogue turn based on context, taking approximately 2.5 hours.

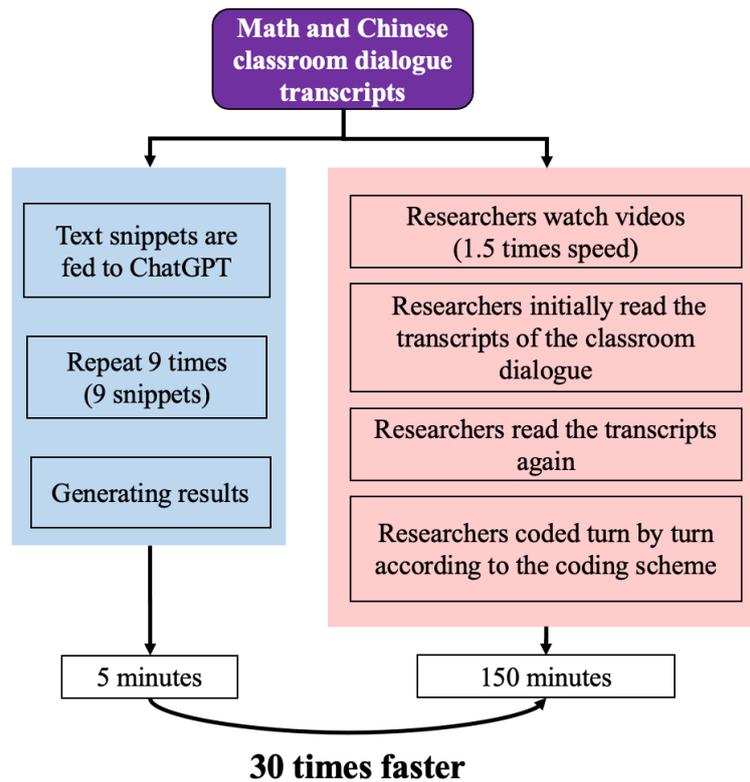

Figure 2. Comparison between manual and automated coding process and time efficiency

Therefore, compared to manual coding, ChatGPT afforded a time-saving factor of approximately 30 times (150 minutes for manual coding versus 5 minutes for ChatGPT). It is noteworthy that this result is based on the performance of an experienced researcher, who minimized the need to frequently consult the coding rules. In scenarios where researchers are less acquainted with the referred coding scheme, manual coding would likely take even longer. This substantial time efficiency offered by ChatGPT emphasizes its potential to significantly expedite the coding process in educational research, aligning with findings from previous studies that highlight the efficiency of AI in data analysis tasks (Kaplan & Haenlein, 2019; Davenport, Guha, Grewal, & Bressgott, 2020).

4.4.2 The consistency evaluation between human coder and ChatGPT

To evaluate the consistency between human researcher and ChatGPT, we conducted intercoder realibility analysis. According to Richards (2009), intercoder reliability "ensures that you yourself are reliably interpreting a code in the same way across time, or that you can rely on your colleagues to use it in the same way." This measure assesses the extent to which independent coders agree upon the application of codes to content, thereby ensuring reliability and validity in qualitative analysis (Lombard,

Snyder-Duch, & Bracken, 2002). For each dialogue turn of analysis, one tallies the number of agreements (where both coders assign the same code) and divides by the total number of dialogue turns to yield a percentage. This figure provides a sense of the coding alignment. However, to adjust for chance agreement, researchers often supplement this with Cohen's Kappa statistic, which offers a more conservative estimate of coder consistency (McHugh, 2012).

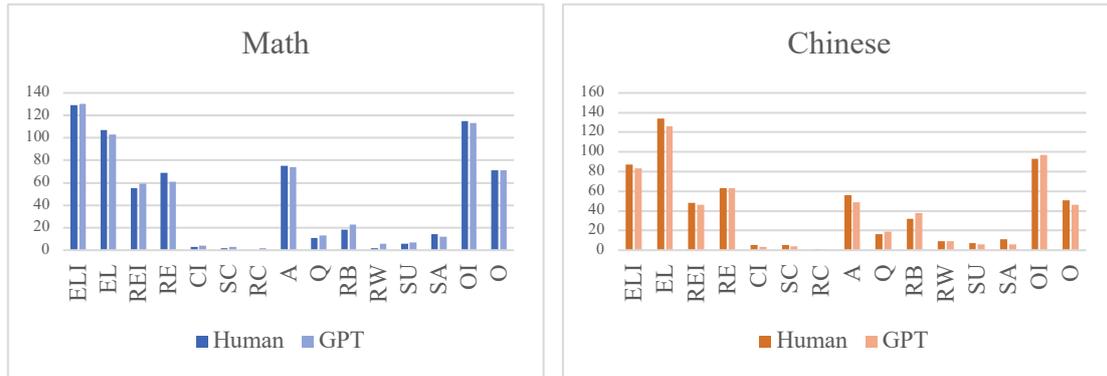

Figure 2. Frequencies of each code analysed by human researcher and ChatGPT

Figure 2 displays the coding outcomes of classroom dialogue analysis in math and Chinese lessons, revealing the frequencies of the 15 codes the researcher and ChatGPT coded. In Math classroom analysis, over six lessons with 576 dialogue turns, the inter-coder agreement percentage between the human coder and ChatGPT is reported at 86.98%. Similarly, in Chinese classroom analysis, which encompasses six lessons with 348 dialogue turns, the agreement percentage is slightly higher at 87.64%. The figure suggests a high level of consistency, indicating that ChatGPT can effectively mirror human coding practices to a significant extent.

Table 2. The Cohen's Kappa statistics of the inter-coder reliability between human coder and ChatGPT

| Codes | Chinese Lessons | Math Lessons |
|---|---|---|
| ELI | 0.973*** | 0.995*** |
| EL | 0.961*** | 0.977*** |
| REI | 0.838*** | 0.962*** |
| RE | 0.947*** | 0.932*** |
| CI | 0.497*** | -0.005 |
| SC | 0.216*** | 0.004 |
| RC | -0.002 | 0[1] |
| A | 0.843*** | 0.992*** |
| Q | 0.735*** | 0.830*** |
| RB | 0.909*** | 0.874*** |
| RW | 0.662*** | -0.004 |

---

[1] No statistics are computed because in human coding results the frequency of RC is 0.

| | | |
|---|---|---|
| SU | 0.611*** | -0.010 |
| SA | 0.702*** | 0.529*** |
| OI | 0.962*** | 0.958*** |
| O | 0.944*** | 0.953*** |

In the context of coding classroom dialogue for educational research, Cohen's Kappa statistic serves as a critical measure for evaluating inter-coder reliability, surpassing mere percentage agreement by accounting for chance agreement (McHugh, 2012). According to Table 2, the Kappa statistics for Chinese and math lessons reveal varied levels of coder concordance. While there is no universally accepted standard for interpreting Cohen's Kappa coefficient, a consensus has emerged in the literature regarding its interpretation. This consensus categorizes the Kappa values as follows: slight (0.00 to 0.20), fair (0.21 to 0.40), moderate (0.41 to 0.60), substantial (0.61 to 0.80), and almost perfect (0.81 to 1.00) (McHugh, 2012; Landis & Koch, 1977). Accordingly, this study adopts these widely recognized benchmarks to analyze the inter-coder reliability presented in Table 2. This approach aligns our analysis with the established norms in the field, ensuring that our interpretations of the Kappa values are both consistent and comparable with existing literature. High Kappa values, as seen in codes such as ELI (0.973 for Chinese, 0.995 for Math) and EL (0.961 for Chinese, 0.977 for Math), indicate near-perfect agreement, underscoring the potential of ChatGPT to accurately follow the coding scheme. Conversely, lower Kappa scores, such as CI (0.497 for Chinese, -0.005 for Math) and SC (0.216 for Chinese, 0.004 for Math), point to a divergence in understanding or application of the coding manual, necessitating further inquiry into these discrepancies. Particularly concerning are the negative Kappa values in RC for Chinese (-0.002) and Math (0.004), which suggest no agreement beyond chance and highlight areas where the AI's coding decisions substantially differ from the human coder. These results prompt a crucial discussion on the refinement of AI coding algorithms for enhanced alignment with human coding practices in educational settings.

## 5 DISCUSSION

The research substantially illustrates the prospects and feasibility of large-scale automated assessment in educational contexts through comprehensive evaluations of time efficiency in manual versus automated coding processes, and calculations of consistency between these methodologies. This examination not only underscores the potential for efficiency improvements but also lays a foundational understanding for the automation of complex tasks. Nevertheless, alongside these promising insights, the study also reveals several points meriting further investigation.

The first issue needs to address is the accuracy issue when comparing ChatGPT's coding of classroom dialogues to human coding. While Cohen's Kappa statistics demonstrate a high level of agreement in codes such as ELI and EL for both Chinese and Math lessons, suggesting that ChatGPT can closely align with the human understanding of the coding scheme, there are notable discrepancies in other codes (McHugh, 2012). These discrepancies, particularly the negative Kappa values seen in codes like RC, underscore the need for ongoing refinement of AI algorithms to enhance their congruence with human coding standards in educational research.

Lower consistency between human researchers and ChatGPT in codes like CI (Coordination Invitation), SC (Simple Coordination), and RC (Reasoned Coordination) can be attributed to differences in contextual interpretation. ChatGPT primarily analyzes textual information and explicit cues, whereas human coders consider broader context and implicit meanings (Hovy & Lavid, 2010). Furthermore, the complexity and subjectivity of these codes impact inter-coder consistency. Coordination, involving multiple interlocutors, requires analyzing various factors and subtle dialogue nuances, challenging for automated coding (Tannen, 1989).

Table 3 exemplifies differing coding results. A human coder might code Turn 4 as CI, based on context, recognizing the teacher's consideration of prior student contributions. Consequently, another student's response in Turn 5 is coded as RC for providing insights with reasons. In contrast, ChatGPT's analysis of Turns 4 and 5 relies more on individual utterances and keywords. Phrases like 'how do you think,' not explicitly inviting elaboration or reasoning, lead to an OI (Other Invitations) code. Similarly, the presence of 'because' in Turn 5 results in a RE (Reasoning) code by ChatGPT.

Table 3. Example of inconsistency in classroom dialogue analysis results between human researcher and GPT

| Turn | Agent | Utterance | Code(s)—Human | Code(s)—GPT |
|---|---|---|---|---|
| 1 | Teacher | What kind of emotions does this poem express? Who would like to analyze it? | OI | OI |
| 2 | Student A | I think the poem mainly expresses admiration for nature. | EL | EL |
| 3 | Student B | I agree with A's view, but I would like to add that there is also a subtle sense of homesickness in the poem. | EL | EL |
| 4 | Teacher | A and B have mentioned admiration for nature and homesickness. How do you think these two emotions are interwoven in the poem? | **CI** | **OI** |
| 5 | Student C | I believe the description of nature brings out the poet's homesickness because the natural scenery reminds the poet of their hometown. | **RC** | **RE** |

Last but not least, the limitation of training data is a significant factor in the consistency disparity between GPT and human analysis in educational dialogues (Jurafsky &

Martin, 2009). Despite ChatGPT's extensive pretraining on vast datasets, the specificity of educational dialogue in the Chinese context requires more targeted data for improved accuracy.

Furthermore, when compared with other models (e.g. Song, 2021), the development of LLMs like ChatGPT marks another encouraging advancement towards the scalability of automatic coding in educational setting. Specifically speaking, previous models had reached approximately 80% inter-coder agreement in a 7-category coding framework, and our model reached more than 85% with a 15-category coding scheme. In other words, these models provide an increasingly accurate automation of coding, paving the way for extensive data analysis that was previously unattainable.

In addition, the dialogue analysis framework employed in this study distinguishes itself by utilizing codes that are fundamentally descriptive of actions, aligning with the principles outlined by Markee (2000) in his comprehensive overview of conversation analysis. This approach marks a significant departure from traditional coding schemes that often rely on underlying theoretical frameworks, such as those discussed by Mercer and Littleton (2007), who explore the role of dialogue in cognitive development from a sociocultural perspective. By focusing on the descriptive nature of the codes, akin to the methodologies described by Ochs, Gonzales, and Jacoby (1996) in their examination of how language mediates classroom interactions, the framework facilitates a theory-neutral stance. This neutrality enhances the precision of automated coding by the large language model, mirroring the action-oriented analysis that simplifies the coding process and boosts accuracy. The subsequent aggregation of these straightforward codes allows for the construction of results that, while initially theory-agnostic, can be interpreted and aligned with theoretical constructs as needed for the research objectives, providing a flexible tool for exploring the complex dynamics of classroom dialogue. Subsequent aggregation of these codes can yield results that are reliant on theoretical constructs, tailored to the specific research needs.

Despite the study's coverage of major subjects like math and Chinese and the inclusion of different contexts of classroom activities such as introduction, practice, and review, the sample size and scope remain limited. A more comprehensive validation across a broader spectrum of disciplines, age groups, and collaborative learning environments is necessary to ensure the generalizability of the findings.

The variation in interaction frequency across different subjects is a significant factor. For instance, research shows that subjects like mathematics often involve more problem-solving and individual work, leading to less frequent but more targeted interactions (Boaler & Staples, 2008). In contrast, subjects like language arts tend to encourage more continuous and exploratory dialogues (Nystrand, & Gamoran, 1991). While in our research, there are less dialogue turns in Chinese lessons comparing to math ones, further investigation on the feature of classroom dialogue in the same subject under different contexts also can be an interesting topic for future research. Age-related differences in student expressiveness also play a crucial role in dialogue and interaction patterns. Younger students may have less sophisticated communication skills, affecting the nature and complexity of classroom dialogues (Mercer & Littleton, 2007). As students mature, their ability to engage in more complex and abstract discussions evolves, leading to a shift in the balance of various dialogue types and coding sequences. Furthermore, the learning environment itself significantly influences interaction modes. Collaborative settings often promote more distributed and peer-focused interactions, contrasting with more teacher-centric dialogues in traditional settings (Gillies, 2004). These variations were not fully represented in our sample.

Therefore, while our study provides valuable insights into classroom interactions, the limitations in sample diversity suggest caution in the broad application of our findings. Future research should aim to include a wider range of subjects, age groups, and learning environments to capture the full spectrum of interaction dynamics in educational settings.

## 6   CONCLUSION

In conclusion, this study underscores the transformative potential of LLMs, particularly GPT-4, in the qualitative analysis of classroom dialogues. The research reveals a significant reduction in the time required for dialogue analysis, demonstrating the efficiency and practicality of AI in educational settings. The high level of consistency between GPT-4 and human coders in most coding categories suggests that AI can effectively mirror human coding practices. This study represents a crucial step towards scalable qualitative educational diagnosis and assessment, offering a promising direction for future research in educational technology. Future work should aim to address the limitations observed, expanding the scope to a broader range of subjects, age groups, and learning environments. This will ensure a more comprehensive understanding and application of AI in educational research, potentially revolutionizing the analysis of classroom interactions and informing teaching practices.